\documentclass{article}

\usepackage{neurips_2022}

\usepackage[utf8]{inputenc} 
\usepackage[T1]{fontenc}    
\usepackage{hyperref}       
\usepackage{url}            
\usepackage{booktabs}       
\usepackage{amsfonts}       
\usepackage{nicefrac}       
\usepackage{microtype}      
\usepackage{xcolor}         
\usepackage{amsmath}
\usepackage{cases}

\usepackage{algorithm}

\usepackage{graphicx}
\usepackage{algorithmicx}
\usepackage{algpseudocode}
\algnewcommand{\LeftComment}[1]{\Statex \(\triangleright\) #1}

\usepackage[inkscapelatex=false]{svg}

\title{
A Secure Aggregation for Federated Learning on Long-Tailed Data 
}

\author{
}

\begin{document}

\maketitle

\begin{abstract}
 As a distributed learning, Federated Learning (FL) faces two challenges: the unbalanced distribution of training data among participants, and the model attack by Byzantine nodes. 
 In this paper, we consider the long-tailed distribution with the presence of Byzantine nodes in the FL scenario.
 A novel two-layer aggregation method is proposed for the rejection of malicious models and the advisable selection of valuable models containing tail class data information. 
 We introduce the concept of think tank to leverage the wisdom of all participants.
 Preliminary experiments validate that the think tank can make effective model selections for global aggregation.
\end{abstract}

\section{Introduction}

Increasing attention has been focused on Federated Learning (FL) with the advancement of machine learning and the trend of decentralized training data.
By cooperatively training models among multiple parties, FL has made great strides in privacy issues and data legislation. 
FL allows participants to share model parameters rather than raw data and update the global model by aggregating local models from the participants. 
However, the distributed training model of FL leads to the issues of security and heterogeneity data distribution among different parties.
Existing global aggregation algorithms in FL, which are based on the global average calculation, are vulnerable to attacks, e.g., FedAvg~[1]. 
The attackers can compromise the accuracy and convergence of the global model by submitting malicious parameters or performing data poisoning attacks.

Existing FL aggregation algorithms, e.g., Krum~[2], median~[3], and trimmed mean~[3], achieve Byzantine Fault Tolerance (BFT) for security by selectively dropping discrepant models trained with distinctive datasets.
The existing global aggregation algorithms in FL overlook the low-frequency and small-size data, which might be of considerable value. 
In the real-world scenario, e.g., Internet of vehicles (IoV) and Internet of Medical Things (IoMT), the obtained data could follow a long-tail distribution with low frequencies and small size because of the imbalanced distribution among FL parties and overall data classes.

A few researchers have focused on the scarce but valuable data resources in imbalanced training data in FL recently.
In~[4], it is first demonstrated that globally imbalanced training data in FL leads to a decrease in model accuracy
To address the problem of declining accuracy, Astraea is developed to rebalance the training process with mediators. 
However, in Astraea, mediators require FL participants to share information about the distribution of their local data, which may introduce new privacy concerns.
The work in~[5], on the other hand, determines whether there is a data imbalance issue in FL by a monitor without directly sharing local data distribution information. 
In each round of FL, the monitor infers the impact of each class on the global model and introduces a new loss function, Ratio Loss, to address the problem of the local imbalance and global imbalance. 
The BalanceFL framework~[6] divides the data imbalance problem into local and global components. 
It uses knowledge inheritance for missing classes (global issue) and balanced sampling for inter-class balancing (local problem), which outperforms the state-of-the-art FL approaches. 
The existing methods focus on the impact of the imbalanced long tail problem on FL accuracy and do not take into account the security issue with the attacks of Byzantine nodes.

In this paper, we propose a novel two-layer aggregation method to fully use the long-tail data with Byzantine nodes. 
We define a new role in the traditional FL process, i.e., the think tank, to discriminate the shared local models for global model optimization. 
The think tank votes on shared models based on their local test results to help the aggregator effectively select models worthy for global aggregation and discard malicious or worthless ones.
In the proposed framework, the aggregation process of FL is divided into two layers, i.e., the filter layer and the vote layer.
In the filter layer, FL aggregators preliminary filter the models by calculating their distances from others.
In the vote layer, the think tank votes on the models dropped by the former and decides whether they should be selected for aggregation to reduce the information loss of the tail classes. 

The think tank, which is composed of all participants, is designed to avoid the problem that models containing information on tail class data are misclassified as anomalies by the existing BFT algorithm like multi-Krum~[2]. 
Compared with other selective aggregation algorithms that compute only on aggregators, the introduction of think tanks can make global decisions smarter and shows the advantages of group cooperation. 
Each participant can be more deeply involved in the FL process by acting as both the provider of the local model and the voters in the think tank.

The main contribution of this paper is that the proposed two-layer aggregation method suppresses malicious updates in FL while preserving models trained from rare samples with an imbalanced long-tail distribution. 
To the best of our knowledge, this paper is the first to consider a combined scenario of imbalanced data processing and Byzantine attack in FL. 
We propose the concept of think tank in FL process to separate the task of judging values of shared local models from aggregators, avoiding the limitations of a single criterion through two-layer validation.

The experimental results show that the proposed two-layer aggregation method improves the accuracy by 9\% over the traditional multi-Krum algorithm when a small amount of unique data exists in a few random nodes in FL.
The proposed method also shows the ability to resist Byzantine node attacks. 


\section{Proposed Method}

The proposed two-layer aggregation method leverages the information of the tail classes data to improve model generalizability while being resilient to attacks, where the training data is long-tail distributed and Byzantine nodes may exist. 
The two layers of the proposed method are the filter layer and vote layer, respectively.
The filter layer filters the underrepresented models based on distance calculation, while the vote layer decides whether the dropped model is beneficial to the global model based on test performance. 
By using the two-stage determination mechanism, the legal models which contain useful knowledge are selected for global aggregation, while the malicious versions are discarded.

 \begin{figure}[!ht]
        \centering
        \includegraphics[width=0.8\columnwidth]{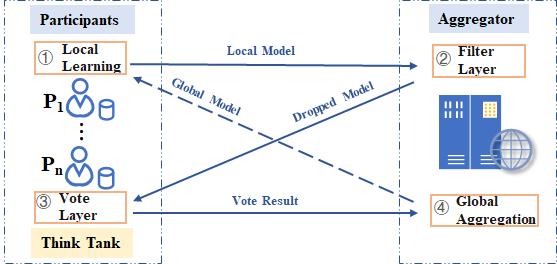}
        \caption{The structure of the proposed method. There are four parts, i.e., local learning, filter layer, vote layer, and global aggregation. FL participants share the locally trained model and participate in the voting layer as the think tank, while the aggregator completes the computation of the filtering layer and the final aggregation.}
        \label{fig_Framework}
\end{figure}

The structure of the proposed method is shown in Fig.~\ref{fig_Framework}, including local learning, filter layer, vote layer, and global aggregation. 
The filter layer and the vote layer are the core of the proposed method to select all valuable models for aggregation while discarding malicious ones. 
Such double standard detection method improves the efficiency of the proposed structure in terms of both information exploitation and attack defense. 
By using the think tank voting process, the wisdom of all participants is utilized for local models’ value evaluation.

Assuming that there are one aggregator and $N$ participants, denoted as $P_1$, $\cdots$, $P_N$, with local dataset $D_i$ ($i\in\{1, \cdots,N\}$), the input of the proposed method in the epoch $r$ is a local model $M_i^r$ shared by a participant to the aggregator, and the output is an updated global model $M_G^r$. 
The two-level process of the proposed aggregation method is described as follows. 

\textbf{Filter Layer.} The aggregator calculates the score $s_i^r$ of $M_i^r$ as given by 
\begin{equation}
    s_i^r = \sum_{k=1}^{K} \| M_i^r - M_k^r \|^2, \qquad  i\in\{1,\cdots,N\},
    \label{equ_equ1}
\end{equation}
where $M_k^r$ refers to the $K$ closest models to $M_i^r$, and the notation $  \| \cdot \|^2 $ indicates the measure of Euclidean distance between models. 
Constant $K$ ($K \leq N$) can be adjusted according to the requirements. 
If model $M_i^r$ is the top $m$ models with the smallest scores, it is selected and we set $I_i^r = 1$. 
Otherwise, we set $I_i^r = 0$. 
According to \eqref{equ_equ2}, the global benchmark model $ M_{G_0}^r$ is calculated. 
For all unselected $ M_{j}^r$, i.e., $I_j^r = 0$, $ M_{j}^r$ is added to the baseline model $ M_{G_0}^r$ for the corresponding hypothetical global model $ M_{G_j}^r$, given by \eqref{equ_equ3}, for further evaluation and vote in the next layer.

\begin{subequations}
\label{eq_ps}
\begin{numcases}{}
    M_{G_0}^{r} = \frac{1}{m}\sum_{i} M_i^r,~\qquad \text{where} \quad I_i^r = 1;
    \label{equ_equ2}
     \\
    M_{G_j}^{r} = \frac{m \times M_{G_0}^{r} + M_{j}^{r}}{m+1},~\qquad \text{where} \quad I_j^r = 0.
    \label{equ_equ3}
\end{numcases}
\end{subequations}

\textbf{Vote Layer.} In the voting layer, the think tank consisting of all participants plays a key role in further evaluating the models filtered in the first layer. 
Each think tanker receives a series of global model candidates and tests the received models on its local test set. 
Think tankers vote on the models not selected in the previous layer based on the test results. 
If $M_{G_j}^{r}$ outperforms $M_{G_0}^{r}$ on $P_i$'s local test set, $P_i$ votes that the model should be added to the global model.

The vote layer evaluates the models based on the accuracy of the test results, which is related to the selection of the test set. 
The test data sets owned by the think tankers can be either assigned from a full test set or randomly sampled from their own local data sets.

The aggregator follows the majority opinion to update the selected results $I_i^r$ of the shared local model $M_i^r$ from each participant $P_i$ based on the votes of the think tanks. 
The final global aggregation result is obtained as given by

\begin{equation}
    M_{G}^{r} = \frac{\sum_{i=1}^{N} M_i^r \times I_i^r }{\sum_{i=1}^{N} I_i^r }.
    \label{equ_equ4}
\end{equation}

Implementation process is detailed in Algorithm~\ref{alg_1}.

\begin{algorithm}[htb] 
\caption{The Proposed Two-layer Aggregation Algorithm.} 
\label{alg_1}
\begin{algorithmic}[1] 

\Require ~~\\ 
Global Epoch $r$, Local Training Model $M_{i}^r$ from Participant $P_i$ ($i=1, \cdots, N$) with Training Dataset $D_i$ and Test Dataset $D_{Test_i}$
\Ensure ~~\\ 
Global model $M_G^r$.

\Statex

\LeftComment{{\textbf{[Filter Layer]}}} 

\State  $score_i^r = \sum \|M_i^{r} - M_k^{r}\|^2$ ($M_k^{r}$ refers to the $K$ closest models to $M_i^r$) 

\If{$score_i^r \in $ [ The $m$ smallest scores]}
    \State  $I_i^r = 1$ 
\Else
    \State  $I_i^r = 0$ 
\EndIf

\State $M_{G_0}^{r} = \frac{\sum_{i=1}^{N} M_i^r \times I_i^r }{m}$ 

\If{$I_j^r = 0$}
    \State $M_{G_j}^{r} = \frac{m \times M_{G_0}^{r} + M_{j}^{r}}{m+1} $ 
\EndIf

\Statex
\LeftComment{{\textbf{[Vote Layer]}}}  

\State  $P_i$ tests $M_{G_0}^{r}$ and $M_{G_j}^{r}$ (for all $j$ that satisfies $I_j^r = 0$) on $D_{Test_i}$ for accuracy $acc_{i,0}^r$ and $acc_{i,j}^r$

\If{$ acc_{i,j}^r \geq acc_{i,0}^r $}
    \State  $t_{i,j}^r = 1$ 
\Else
    \State  $t_{i,j}^r = 0$
\EndIf

\If{$ \sum_{i=1}^{N} t_{i,j}^r \geq \frac{N}{2}$ and $I_j^r = 0$}
    \State $I_j^r = 1$ 
\EndIf

\Statex
\LeftComment{{\textbf{[Global Aggregation]}}}
\State $M_{G}^{r} = \frac{\sum_{i=1}^{N} M_i^r \times I_i^r }{\sum_{i=1}^{N} I_i^r } $ 

\Statex

\Return{$M_G^r$}
\end{algorithmic}
\end{algorithm}

\section{Experiment Plan and Progress}

\subsection{Experiment Plan} 
The overall experiments include two parts: the classification accuracy under the different data settings and the different number of Byzantine nodes.

First, the experiments for different data settings can be further divided into data set selection, training data distribution, and test dataset settings.
The datasets will use artificially imbalanced MNIST datasets and ImageNet-LT~[7]. 
By artificially controlling the ratio of each class in the MNIST dataset, we will draw a clear picture of the performance of the provided method under different training data distribution parameters, such as the ratio of the number of target classes to other classes and the frequency of the target classes appearing among participants. 
With ImageNet-LT, which simulates the actual long-tail data in nature, we want to show the value of the proposed method in practical applications. 

As the test sets used by the think tank also have a significant impact on the aggregation results, a series of experiments are also needed to show the effects of the different test sets on the proposed method, for example, test sets of the think tank contain full class data or only have imbalanced test data consistent with the training data.

As for the research of Byzantine nodes, complete and systematic experiments will be conducted to compare and demonstrate the resilience of the proposed method under different Byzantine node settings.
The method that can provide a larger number of Byzantine fault-tolerant nodes is considered to be superior. 
We start our experiments with a small number of Byzantine nodes.

\subsection{Current Progress} 
From the present experimental results, the proposed method is able to learn rare training sample knowledge effectively and free from attacks. 
We simulate the situation of the tail class on MNIST by deliberately setting the training data of class 0 to exist only on 10\% participants and to have only 10\% the amount of data of the other classes. 
The test set used by each think tanker is randomly sampled from a complete test set with a balanced distribution of classes, as the test criteria require equally good generalization to a small and less frequent class. 
In addition, we set up a Byzantine node with 10\% class 0 data but maliciously reverses its label to the wrong one to confuse the aggregator.

\begin{figure}[!ht]
    \begin{minipage}[t]{0.48\textwidth}
        \centering
        \includegraphics[width=1\columnwidth]{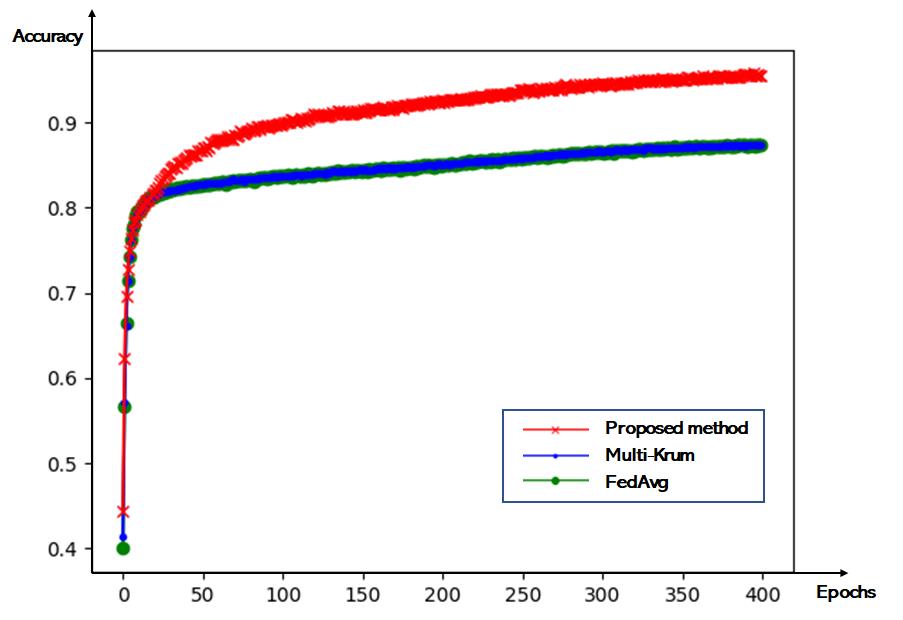}
        \caption{Test accuracy on the full test set. Only one participant in ten has a small amount of class~0 data (10\% of the other classes).}
        \label{fig_accuracy}
    \end{minipage}
    \hspace{4mm}
    \begin{minipage}[t]{0.48\textwidth}
        \centering
        \includegraphics[width=1\columnwidth]{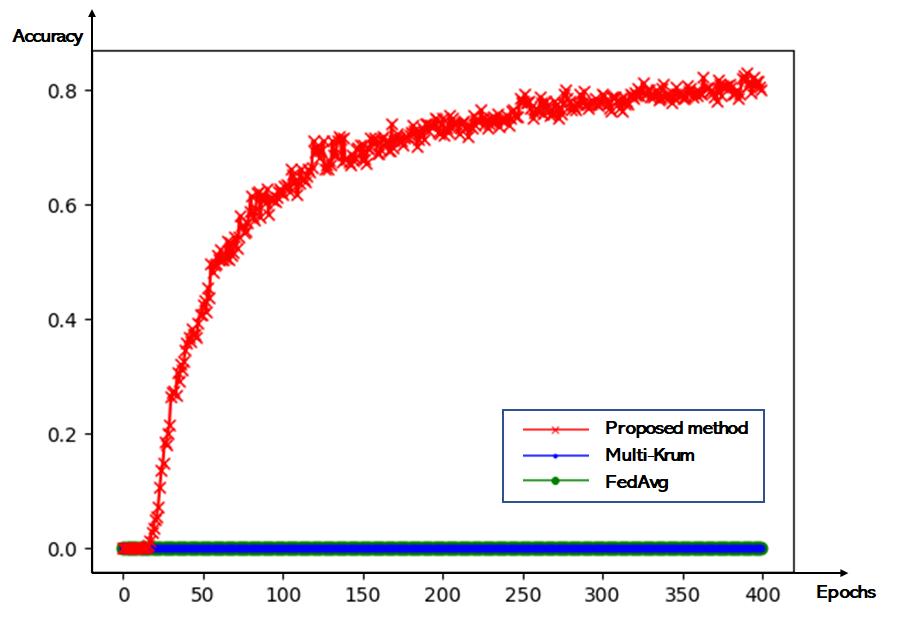}
        \caption{Test accuracy on the class 0 test set. Only one participant in ten has a small amount of class~0 data (10\% of the other classes).}
        \label{fig_accuracy_0}
    \end{minipage}
\end{figure}

We compare the proposed method with FedAvg~[1] and multi-Krum~[2] algorithms. 
Fig.~\ref{fig_accuracy} and Fig.~\ref{fig_accuracy_0} show the accuracy of the three methods on the complete test set and on a separate test set consisting of a particular class, class 0, respectively.

We see that both FedAvg and multi-Krum algorithms fail to learn recognition on class 0 data with similar classification accuracy, because the FedAvg algorithm is affected by the malicious model and the multi-Krum algorithm directly discards the model containing class 0 data as well as the malicious one. 
The proposed model performs better than its counterparts in the same scenario that it can effectively learn the knowledge of class 0 data with limited information. 
The proposed model improves the accuracy of the model on the overall test set from 87.29\% for multi-Krum and 87.44\% for FedAvg to 95.51\%, an improvement of 9.42\% and 9.23\%, respectively. 
The accuracy on the class 0 data is enhanced from up to 81.6\% by the proposed method. 
Furthermore, after a short period (about 20 epochs), the proposed method can accurately reject malicious models during aggregation to protect the models from attacks.

\section{Discussion and Future Work}


In this paper, we present a two-layer aggregation method as a safe solution to the long-tail data issue in FL with Byzantine nodes, which can identify malicious attacks and learn useful knowledge from the small amount and low-frequency tail class training data. 
The think tank made up of FL participants is designed to help aggregators more effectively and accurately measure the value of shared local models.

For the maximum fault-tolerant node number of the proposed method, we will use mathematical methods for theoretical derivation. 
A more complete and rigorous proof of the BFT property of the proposed method will be given in conjunction with the corresponding experimental results. 
In addition, considering the computational and communication costs, we will perform accurate calculations by mathematical methods and demonstrate them by experiments.



\newpage


\vspace{10em}

\section*{References}

\medskip

{
\small

[1] McMahan, B., Moore, E., Ramage, D., Hampson, S., \& y Arcas, B. A. (2017, April). Communication-efficient learning of deep networks from decentralized data. In Artificial intelligence and statistics (pp. 1273-1282). PMLR. 

[2] Blanchard, P., El Mhamdi, E. M., Guerraoui, R., \& Stainer, J. (2017). Machine learning with adversaries: Byzantine tolerant gradient descent. Advances in Neural Information Processing Systems, 30.

[3] Yin, D., Chen, Y., Kannan, R., \& Bartlett, P. (2018, July). Byzantine-robust distributed learning: Towards optimal statistical rates. In International Conference on Machine Learning (pp. 5650-5659). PMLR.





[4] Duan, M., Liu, D., Chen, X., Tan, Y., Ren, J., Qiao, L., \& Liang, L. (2019, November). Astraea: Self-balancing federated learning for improving classification accuracy of mobile deep learning applications. In 2019 IEEE 37th international conference on computer design (ICCD) (pp. 246-254). IEEE.

[5] Wang, L., Xu, S., Wang, X., \& Zhu, Q. (2021, May). Addressing class imbalance in federated learning. In Proceedings of the AAAI Conference on Artificial Intelligence (Vol. 35, No. 11, pp. 10165-10173).

[6] Shuai, X., Shen, Y., Jiang, S., Zhao, Z., Yan, Z., \& Xing, G. (2022, May). BalanceFL: Addressing Class Imbalance in Long-Tail Federated Learning. In 2022 21st ACM/IEEE International Conference on Information Processing in Sensor Networks (IPSN) (pp. 271-284). IEEE.

[7] Liu, Z., Miao, Z., Zhan, X., Wang, J., Gong, B., \& Yu, S. X. (2019). Large-scale long-tailed recognition in an open world. In Proceedings of the IEEE/CVF Conference on Computer Vision and Pattern Recognition (pp. 2537-2546).

}

\end{document}